\documentclass[journal]{IEEEtran}
\usepackage{amsmath}
\usepackage{bbm}
\usepackage{mathtools}
\usepackage{breqn}
\usepackage{multicol}
\usepackage{soul}
\usepackage{microtype}
\usepackage{graphicx}
\usepackage{booktabs} % for professional tables
\usepackage{amsmath}
\usepackage{amsfonts}
\usepackage{wrapfig}
\usepackage[table]{xcolor}
\usepackage{dirtytalk}
\usepackage{algorithm} 
\usepackage{algpseudocode} 
\usepackage{listings} 
\DeclareMathOperator*{\argmax}{argmax}
\usepackage{listings}
\usepackage{framed}
\DeclarePairedDelimiter\floor{\lfloor}{\rfloor}

\begin{document}
\title{CycleCluster: Modernising Clustering Regularisation for Deep Semi-Supervised Classification}

\author{Philip~Sellars$^1$, Angelica I. Aviles-Rivero$^1$ and Carola-Bibiane Sch{\"o}nlieb$^1$%
\thanks{P. Sellars, Angelica I. Aviles-Rivero and  Carola-Bibiane Sch{\"o}nlieb are with the Department of Theoretical Physics and Applied Mathematics, Univeristy of Cambridge, Cambridge, UK.  {ps644,ai323,cbs31}@cam.ac.uk .}
}

\maketitle

\begin{abstract}
Given the potential difficulties in obtaining large quantities of labelled data, many works have explored the use of deep semi-supervised learning, which uses both labelled and unlabelled data to train a neural network architecture. The vast majority of SSL approaches focus on implementing the \textit{low-density separation assumption} or \textit{consistency assumption}, the idea that decision boundaries should lie in low density regions. However, they have implemented this assumption by making local changes to the decision boundary at each data point, ignoring the global structure of the data. In this work, we explore an alternative approach using the global information present in the clustered data to update our decision boundaries. We propose a novel framework, CycleCluster, for deep semi-supervised classification. Our core optimisation is driven by a new clustering based  regularisation along with a graph based pseudo-labels and a shared deep network. Demonstrating that direct implementation of the \textit{cluster assumption} is a viable alternative to the popular consistency based regularisation. We demonstrate the predictive capability of our technique through a careful set of numerical results.  
\end{abstract}

\section{Introduction}
Deep Learning (DL) has achieved state-of-the-art results in many different task including object detection e.g.~\cite{ren2015faster,girshick2015fast,redmon2016you},  segmentation e.g.~\cite{long2015fully, ronneberger2015u, chen2017deeplab}, deraining e.g.~\cite{yang2019joint,Li_2019_ICCV} and classification e.g.~\cite{krizhevsky2012imagenet,he2016deep,hu2018squeeze}. The core assumption of these supervised approaches is that they rely upon a large, accurate and representative dataset to allow for good generalisation to unseen examples. However, in real-world applications obtaining annotations are time consuming, expensive and can require expert knowledge in technical domains. This has motivated the fast development of techniques that can exploit unlabelled data \cite{Ji_2019_ICCV,Mahasseni_2017_CVPR}.

The community has reported promising results in using SSL for image classification, in which the vast majority of approaches are \textit{consistency-enforcing} approaches including~\cite{bachman2014learning, tarvainen2017mean,  miyato2018virtual, verma2019interpolation}.  That is, they follow the key assumptions that allow SSL to work ~\cite{chapelle2003cluster,chapelle2009semi}:  i) close points are likely to have the same label (i.e. the so-called \textit{smoothness assumption}), and  ii) decision boundaries should lie in low density regions (i.e. the so-called \textit{low-density separation assumption}).  The second assumption can be seen as a special case of the first . The low-density assumption is equivalent to the cluster assumption, points in the same cluster are likely to be in the same class \cite{chapelle2003cluster,chapelle2009semi}. However, whilst many works have implemented the low-density assumption by adding a domain specific perturbation factor $\delta$ to the unlabelled data or weights and enforcing invariant predictions with respect to $\delta$, no one has investigated the impact of directly implementing the cluster assumption. 

However, the question of how to set $\delta$ is not trivial and relying on random perturbations to form a representative search of the local feature space becomes computational infeasible in high dimensions. There are several works that have addressed this difficulty - for example using Generative Adversarial Nets (GANs) e.g.~\cite{miyato2016adversarial,miyato2018virtual} to learn $\delta$ or interpolation e.g.~\cite{verma2019interpolation} which limits $\delta$ to be transformations between unlabelled data points.  These alternatives have reported great results on SSL. However, they are also limited by their own construction; for example, it has been recently shown that adversarial training can limit the generalisation capabilities in SSL approaches~\cite{nakkiran2019adversarial}. However, more fundamentally, these methods treat datasets as a set of single entities, where the impact of $\delta$ is designed to affect the feature space around each entity separately. They discount relevant assumptions about SSL such as the strong relationship between entities. The ideal $\delta$ for a point $x_i$ should be dependent on the distribution of the dataset at $x_i$. 

{\textbf{Contributions.}} In this work, we present a \textit{novel general alternative} to the domain specific $\delta$-based approaches based around direct implementation of the cluster assumption. We propose a new approach, which we term CycleCluster, based around simultaneously training a shared architecture on a unsupervised cluster based task and a semi-supervised pseudo-label task. Using the cluster assumption we are able to use global information from the unlabelled dataset to learn better decision boundaries which then allows for the generation of more meaningful pseudo-labels. Our modelling hypothesis is that by carefully combining our clustering regularisation approach to pseudo-label approaches we can greatly boost performance. We demonstrate through rigorous experiments on benchmark datasets that this is the case and that \textit{clustering regularisation is a strong viable alternative to $\delta$-perturbation techniques}. Furthermore, we perform cluster based ablation experiments and show that the common problem of choosing the number of clusters is not a problem in our framework.

\section{Related Work}
The application of SSL has been widely investigated since the early developments in the area e.g.~\cite{zhou2004learning,zhu2003semi,kim2009semi}. With the advent of deep learning, many methods have applied deep learning to the task of semi-supervised learning. In this related work we first visit the topic of consistency regularisation and graphical pseudo-labels for deep neural networks before exploring the task of cluster based learning. 

\subsection{Consistency Regularisation}
Several Deep SSL SOTA-models are based on consistency regularisation, in which the main idea is that an induced perturbation $\delta$ on the data input shall not change the value of the output $f(x)$, so that $f(x) = f(x +\delta)$. This condition can be applied to both the labelled and unlabelled data points. Within this philosophy several current works have been proposed.

The $\prod-$model~\cite{laine2016temporal} is based on inducing stochastic perturbations, in which output consistency is enforced by evaluating each unlabeled sample twice in the network. The output is then computed by minimising the difference in class probability between the two realisations. In the same work, authors introduced the Temporal Ensembling~\cite{laine2016temporal} model. It simplifies the previous model by considering the network predictions over several previous epochs. The  $\prod-$model is an special case of the work of Sajjadi et al ~\cite{sajjadi2016regularization}, and a simplification of the $\Gamma-$model~\cite{rasmus2015semi}.  

Although Temporal Ensembling~\cite{laine2016temporal} was an improvement over previous models, it has a major drawback in that its targets are only updated once per epoch, which bottlenecks the transfer of the learned information to the training process. 
To mitigate this problem, and what might be the current top reference for deep SSL, Tarvainen \& Valpola proposed the Mean Teacher~\cite{tarvainen2017mean} model. The central idea is to maintain an exponential moving average of the network parameters rather than average label predictions. 

Following a philosophy close to $\prod-$model, Virtual Adversarial Training (VAT)~\cite{miyato2018virtual} proposed using adversarial perturbations to measure the local smoothness of the input. They based this approach on the sense of relating distributional divergence to the $\delta$ that maximises the change of the output prediction . The VAT approach has served as complement to other approaches. For example, the work of that~\cite{park2018adversarial} which introduces adversarial dropout, in which the divergence term enforces more robust predictions. More recently, the authors of \cite{jackson2019semi} proposed an approach that seeks to map points into the model parameter space. This is then used to minimise the distance between the label and unlabelled data. 

Another set of techniques report state-of-the-art results e.g.~\cite{berthelot2019mixmatch,sohn2020fixmatch,xie2020unsupervised}, whilst relying on strong augmentations along with complex optimisations schemes . However, these methods are strongly reliant on strong augmentations and diverge when the augmentation stratergy is changed and it is unclear to what extent the performance depends on the methods versus the augmentation choice.
With this motivation in mind, we initially limit our comparison to methods that only use weak perturbations, and not strong data augmentation techniques, to fairly compare the effect of clustering regularisation to $\delta$-perturbation techniques. We then demonstrate how augmentation can be combine with our approach to boost performance and provide initial results.

As an alternative, one can exploit the rich structure of a graph to improve predictions. The top reference method for graph based SSL is Label Propagation~\cite{zhu2005semi} (LP), whose performance heavily relies upon the initial construction of the graph. Most recent works have push the limits of LP by introducing learnt feature information to construct the graph including~\cite{yang2016revisiting,aviles2019beyond,liu2019deep}. Most recently and closely related work to our work, Iscen et al.~\cite{iscen2019label} scaled the classical work of for Zhou to deep networks.

\subsection{Clustering Task}
We also mention the closely related problem of clustering. The central idea is to partition a given dataset into multiple clusters, with maximal inter-cluster similarity and minimal intra-cluster distance. This problem has been widely explored in the literature, and in the field of deep learning including works of that~\cite{coates2012learning,yang2016joint,bojanowski2017unsupervised}. Recently, in the work of Caron et al.~\cite{caron2018deep}, the authors proposed a scalable clustering approach that alternates between the popular k-means algorithm and the updating the parameters of a deep learning network. In semi-supervised learning Margin-Mix \cite{marginmix}), use a class margin loss to encourage each class to cluster together and apart from other classes. These class centroids are then used to produce class pseudo-labels. This approach is vastly different to ours as we use a truly unsupervised clustering algorithm, rather than using a class margin loss, which allows us to use an arbitrary number of clusters unlike Margin-Mix where the number of clusters must be the number of classes. Furthermore, our clustering task produces clustering pseudo-labels which are unrelated to the classification problem.  

Our hypothesis is that, and unlike existing works relying solely on consistency regularisation, the explicit implementation of the clustering assumption can boost the genearlisation of the network. To achieve this, our work is inspired by the principles of \textit{deep unsupervised feature learning} \cite{coates2012learning}. 
Pseudo-label approaches often have a generalisation bottleneck as the initial feature representation is heavily dependent upon the few initial labels. Consistency regularisation investigates points, $f(x+\delta)$, close to the original labels which contains a high amount of mutual information. Instead, we propose using a clustering based approach to learn the global structure of the dataset, improving the feature representation and thus providing better pseudo-labels. 

\begin{figure*}[t!]
\begin{centering}
\includegraphics[width=1\linewidth]{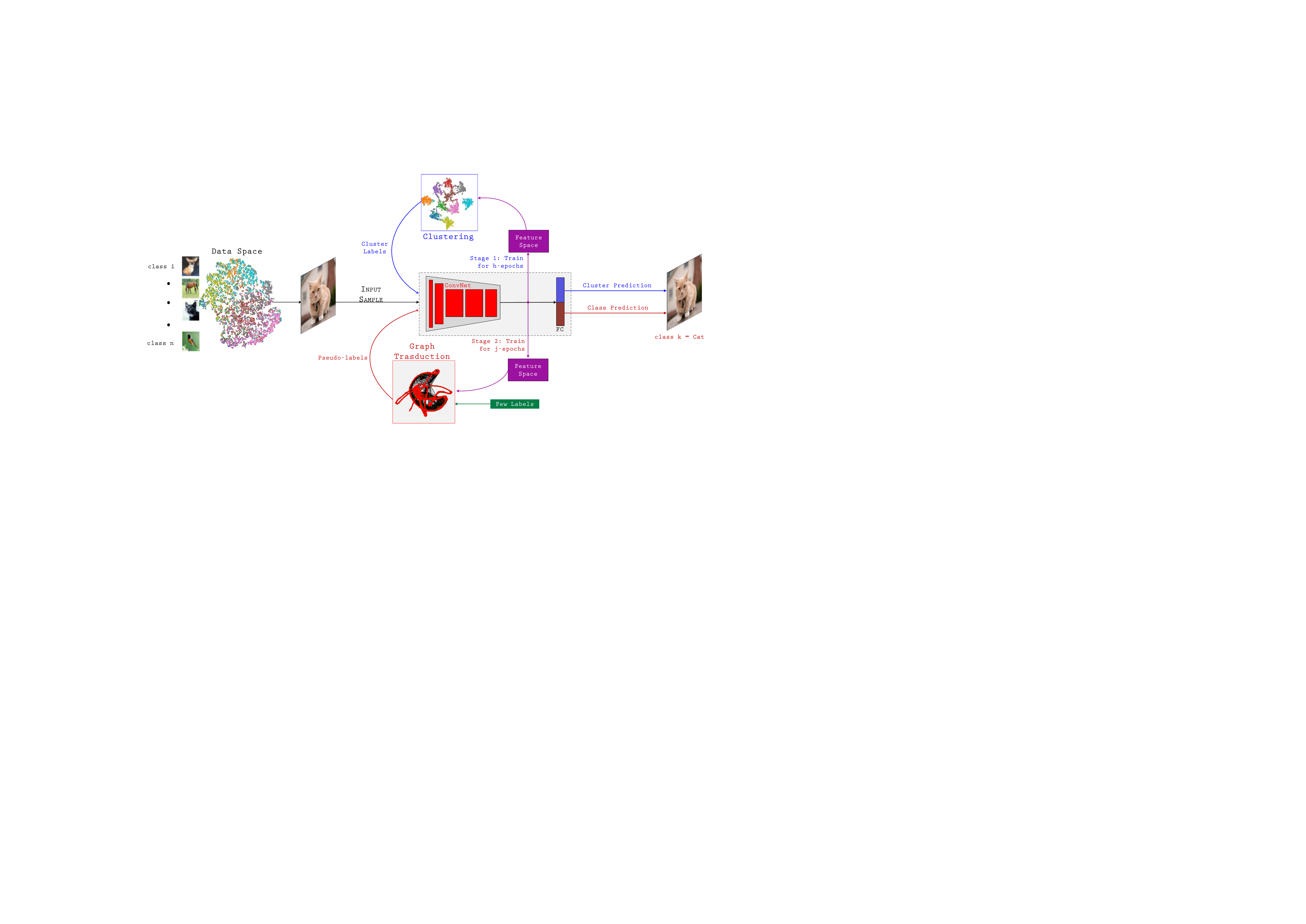}
\par\end{centering}
\caption{Our approach consists of two separate tasks that share the same architecture. Firstly data is fed into the network and the feature representation is extracted, shown in purple. Using these features we then perform two different methods of \textit{pseudo-label} generation. Using the top cycle, in blue, we cluster the feature space and output the cluster assignments as \textit{unsupervised pseudo-labels}, which also acts as the \textit{cluster assumption}. Using the lower cycle, we construct a graphical representation of our data before diffusing the initial labels to generate \textit{semi-supervised pseudo-labels}, which also acts as our \textit{smoothness assumption}. Using the two different sets of pseudo-labels we train the network to predict both the clusters and class of each data point. Note that the same FC layer is used for both tasks. If the number of clusters $K$ is greater than the number of classes $C$, then class prediction is given by the output of neurons $\{1,..,C\}$.}
\label{fig:illustration}
\end{figure*}

\section{Proposed Approach}
In this section, we introduce our novel semi-supervised learning approach that builds on the clustering and smoothness assumptions.  In what follows, we detail each part and start by explicitly defining the problem at hand.

\textbf{Problem Statement.} From a joint distribution $\mathcal{Z} = (\mathcal{X},\mathcal{Y})$ we have a dataset $Z$ of size $n = n_l + n_u$ comprised of a labelled part of joint samples $Z_l= \{ x_i, y_i \}_{i=1}^{n_l}$ and a unlabelled part $Z_u = \{ x_i \}_{i=n_l+1}^{n}$ of single samples on $\mathcal{X}$. The labels come from a discrete set of size C $y \in \{1,2,..,C\}$. Our task is to train a classifier $f_{\theta}$, modelled by a neural network with parameter vector $\theta$, which can accurately predict the labels of unseen data samples from the same distribution $\mathcal{X}$.  The classifier $f$  can be viewed as the composition of two functions $z$ and $g$ such that $f_{\theta}(x) = g_{\theta}(z_{\theta}(x))$. $z_{\theta} : \mathcal{X} \rightarrow \mathbb{R}^{d_p}$ is the embedding function mapping our data input to some $d_p$ dimensional feature space and $g_{\theta} : \mathbb{R}^{d_p} \rightarrow \mathbb{R}^{C}$ projects from the feature space to the classification space. 

We address this problem by proposing a novel framework that alternates between two learning tasks on one shared neural architecture and provide Figure \ref{fig:illustration} as a visual guide. Our first task is a cluster regularisation that pushes decision boundaries to low-density regions in a global sense, and our second is pseudo-label learning from a transductive graph based approach that then can benefit from the better feature representation. 

\subsection{Clustering Regularisation}

\begin{figure}[h!]
\begin{centering}
\includegraphics[height=6cm]{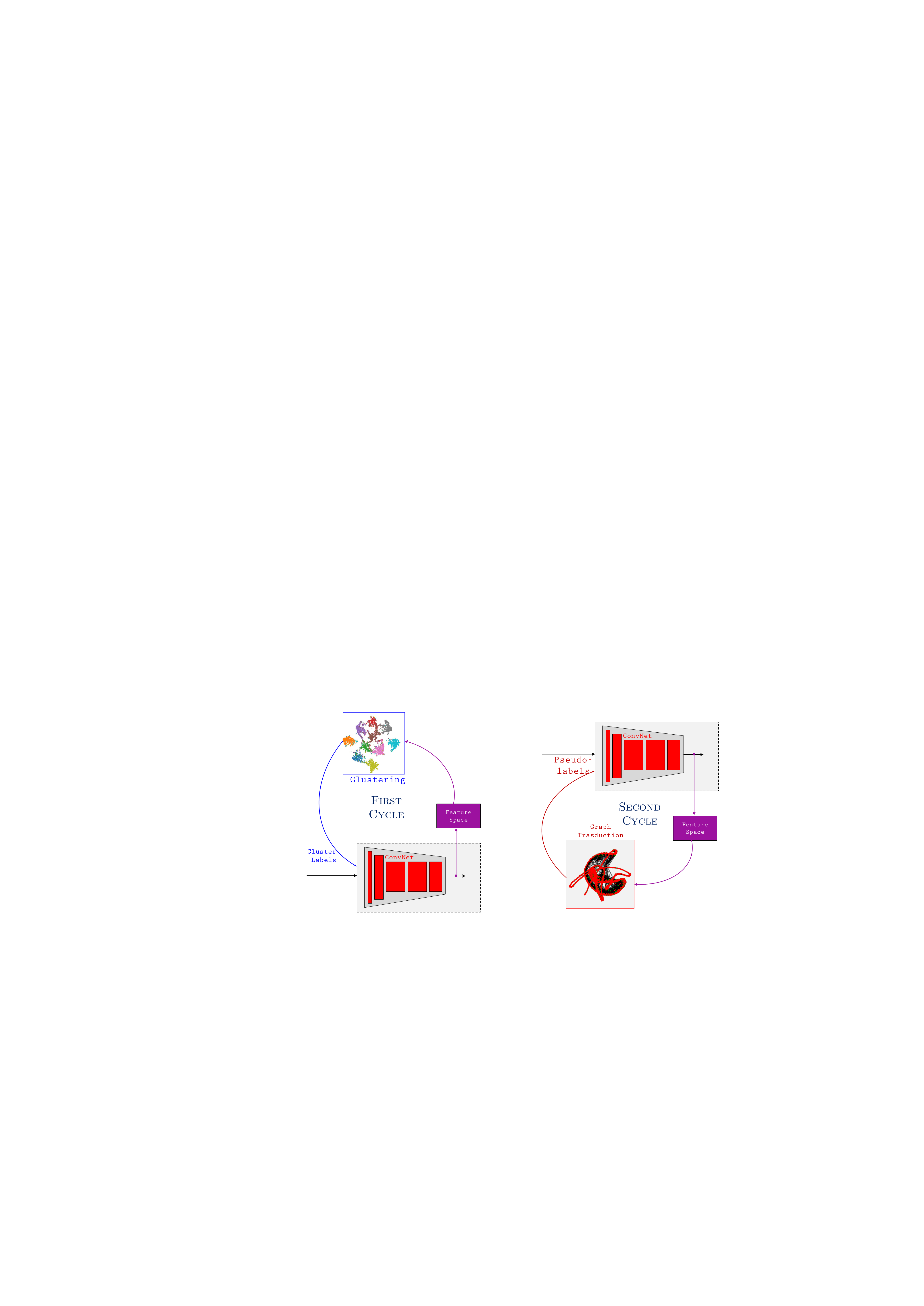}
\par\end{centering}
\caption{Cluster based pseudo-label extraction.}
\label{fig:firstCycle}
\end{figure}

In this paper, we revert back to the original \textit{clustering assumption} of SSL \cite{SSTheory}, which motivates our first learning problem - see Fig.~\ref{fig:firstCycle}. In that we assume points in the same cluster are likely to share the same label. The majority of SOTA-models take the equivalent assumption termed  \textit{low-density separation}. Instead, in this work, we argue that by carefully considering the original clustering assumption one can boost overall performance past the level of low-density separation approaches.

With this approach we need to first cluster our data and then extract meaningful labels from which we can train our neural network. In order to cluster large-scale datasets we need a fast yet powerful clustering algorithm. One of the most popular algorithms is Lloyd's $K-$means \cite{lloyd1982least} algorithm and that is what we use in our approach. Many SOTA models for unsupervised learning, including those based in deep learning e.g.~\cite{caron2018deep}, build upon it.  However, a major drawback is setting the number of clusters $k$ but we show that this problem can be easily managed in our framework. We take inspiration from a observation in~\cite{ren2003learning,achanta2012slic}  \textit{over-segmentation increases discriminative information} which have been demonstrate again recently for big datasets such as in~\cite{caron2018deep,sellars2019superpixel}. However, the benefits of over-clustering have not been investigated for semi-supervised learning.

More precisely, given input data  $X = \{x_n\}_{n=1}^{n}$, we seek to partition $X$ into $K$ clusters. Each cluster is characterised by a centroid. We take $z_{\theta}(X)$ as the feature representation of our input and seek to solve a joint optimisation over the centroid matrix $\mathcal{M}\in \mathbb{R}^{d_p \times K}$ and the clusters assignments $\tilde{Y} = \{ \tilde{y}_1 , .. ,\tilde{y}_n\}$ where $\tilde{y}_I\in\{0,1\}^k$. We then use the cluster assignments $\tilde{Y}$ as unsupervised pseudo-labels and train the network to predict the clusters assignments. To this end, we seek to solve the following loss:

\begin{equation}
   \theta \leftarrow L_C(X,\tilde{Y};\theta) := \frac{1}{n}\sum_{i=1}^{n} l_s(f_\theta(x_i),\tilde{y}_i),  \label{clus-loss}
\end{equation}

\noindent
In this paper, we use \textit{cross-entropy} as the loss function. We can use this loss straight from model initialisation as the performance of even randomly initialised ConvNets on standard transfer tasks, is far above the performance of chance. This is linked to the strong prior that the convolutional architecture puts upon the data.

\subsection{Graph-Based Pseudo-Labels}
In this section, we discuss our second learning task, semi-supervised learning with pseudo-labels and how it is connects to the cluster regularisation. As well as the clustering assumption, the ability for SSL to yield increases in performance also relies on the \textit{smoothness assumption}, in that if two points $x_1,x_2$ are close then the corresponding outputs $y_1,y_2$ should also be close. \textit{In the context of neural networks we rewrite this as}, if two feature representations $z_{\theta}(x_1),z_{\theta}(x_2)$ are close then their outputs $y_1,y_2$ should be close to. To enforce this constraint we use the classical label propagation (LP) approach ~\cite{zhu2002learning} to predict the labels for unlabeled data points and then used these to train our network. In this work, we use the approach of Zhou et al~\cite{zhou2004learning} as the backbone of our method and give a brief overview here. 

With LP we can generate pseudo-labels $\hat{y}_i$ for each unlabelled example $x_i$. \textbf{How do we do this?} We take a dataset $X = X_L + X_U$ and labels $Y_L$ to  construct a weighted graphical representation of the data. From there we can propagate the initial label information over the graph by minimising the graphical Laplacian functional and a label fidelity term and obtaining the prediction matrix $F$ as

\begin{equation}
    \mathcal{Q}(F) = \frac{1}{2}   \sum_{i,j =1}^{n} \mathcal{W}_{ij} \left|\left| \frac{F_i}{\sqrt{D_{ii}}} - \frac{F_j}{\sqrt{D_{jj}}} \right|\right|^2
    + \frac{\mu}{2} \sum_{i=1}^{n} || F_i - Y_i ||^{2} ,
    \label{graph:cost}
\end{equation}

\noindent
where $W$ is the normalised weighted adjacency matrix, $D$ is the degree matrix, $Y$ is the initial label information and $\mu$ is a balancing parameter. From this we extract the pseudo-labels by taking the row maximum $\hat{y}_i = \argmax_j F_{ij}$. We can then train our model on the pseudo-labels $\hat{Y}_U := (\hat{y}_{n_l+1},...,\hat{y}_n )$ for the unlabelled data samples $Z_U$. 

To combat the problems of pseudo-label certainty and class balancing we use a \textit{class weight} $\zeta_{y_i} \in (0,1)$ to account for unbalanced pseudo-labels and to account for pseudo-label uncertainty we use the approach suggest by Iscen et al \cite{iscen2019label} and include an \textit{entropy weight} $w_i \in (0,1)$ which encodes the certainty of an individual label. Higher entropy pseudo-labels are weighted less favourably compared to lower entropy pseudo-labels. Then our loss function, over both the labelled and unlabelled data points, reads:

\begin{equation}
\begin{aligned}
L_W(X_u,Y_L,\hat{Y}_U;\theta) :=  \frac{1}{n_l} \sum_{i=1}^{n_l} \zeta_{y_i} l_s (f_{\theta}(x_i),\hat{y}_i) \\
+ \frac{1}{n-n_l}\sum_{i=n_l+1}^{n}\zeta_{y_i} \omega_i l_s (f_{\theta}(x_i),\hat{y}_i)
\end{aligned}
\tag{6}
\end{equation}

\subsection{Cyclic Optimisation}
We combine the optimisation of these two tasks on the same shared framework to simultaneously exploit the clustering and pseudo label generation tasks. We do so in the following way. At the start of each epoch we extract the feature representation of the data and extract the cluster pseudo-labels via $K$-means clustering and label propagation. From this we optimise $L_C(X,\tilde{Y},\theta)$ for one pass through the whole dataset $Z$ before optimising $L_{W}$ for one pass through the unlabelled data. Therefore the labels are produced once at the start of each epoch prior to the parameter updates. The reason for this choice of cyclical rather than joint loss was that the clustering task produces a good feature representation but the clustering task differs from a classification task. Therefore, the semi-supervised classification is used to tune the model to the task at hand. To give further clarity on our methods we provide a full algorithm in Section 1 in the supplementary material.

\begin{table*}[h!]
\begin{centering}
\resizebox{0.9\textwidth}{!}{ 
\begin{tabular}{|ccccc|}
\hline
\multicolumn{5}{|c|}{\cellcolor[HTML]{D0D0D0} \textsc{CIFAR-10}}  \\ \hline
\multicolumn{1}{|c|}{\textsc{}} & \multicolumn{4}{c|}{\#  \textsc{Labels}} \\ \cline{1-5}
\multicolumn{1}{|c|}{\textsc{Method}}  &  500 & 1k & 2k & 4k   \\ \hline
\multicolumn{1}{|c|}{Fully Supervised}   & 48.93$\pm$0.80 & 39.18$\pm$0.88 & 28.23$\pm$0.49 & 21.20$\pm$0.46   \\\hline\hline
\multicolumn{1}{|c|}{Ladder Networks~\cite{rasmus2015semi}} &  $-$ &    $-$ & $-$ & 20.40$\pm$0.47  \\
\multicolumn{1}{|c|}{VAT \cite{miyato2018virtual} }   & $-$ &  $-$ & $-$ & 11.36$\pm$0.34  \\
\multicolumn{1}{|c|}{SSL-GAN~\cite{salimans2016improved} } & $-$  & 21.83$\pm$2.01 & 19.61$\pm$2.09 & 18.63$\pm$2.32          \\
\multicolumn{1}{|c|}{TSSDL~\cite{shi2018transductive}  $\dagger$} & $-$  & 21.13$\pm$ 1.17 & 14.65$\pm$ 0.33 & 10.90 $\pm$ 0.23        \\
\multicolumn{1}{|c|}{MT~\cite{tarvainen2017mean}}      &  27.45 $\pm$ 2.64    & 21.55$\pm$1.48&	15.73$\pm$0.31 &	12.31$\pm$0.2         \\
\multicolumn{1}{|c|}{ICT ~\cite{verma2019interpolation}  $\dagger$} &  $-$ &     19.56$\pm$0.56 &	14.35$\pm$0.15	& 11.19$\pm$0.14      \\ 
\multicolumn{1}{|c|}{LPDSSL~\cite{iscen2019label}  $\dagger$}  & 32.40 $\pm$ 1.80  & 22.02 $\pm$ 0.88   &	15.66$\pm$0.35 & 12.69$\pm$0.29   \\ 
\multicolumn{1}{|c|}{LPDSSL + MT~\cite{iscen2019label}  $\dagger$}  & 24.02 $\pm$ 2.44  & 16.93 $\pm$ 0.70   &	13.22$\pm$0.29\% & 10.61$\pm$0.28   \\

\multicolumn{1}{|c|}{LGA ~\cite{jackson2019semi}  $\dagger$} &  $-$ &     $-$ &	$-$	& 12.91$\pm$0.15     \\ 
\multicolumn{1}{|c|}{LGA + VAT~\cite{salimans2016improved} $\dagger$ } & $-$  & $-$ & $-$ & 12.06 $\pm$ 0.19\\ \hline\hline

\multicolumn{1}{|c|}{CycleCluster}  & \textbf{19.35 $\pm$ 2.52} & \textbf{14.76$\pm$ 0.34} & \textbf{ 12.11 $\pm$ 0.40 } & \textbf{10.52 $\pm$ 0.45}\\  \hline 
\end{tabular} 
}
\caption{Comparison with SSL methods on CIFAR-10. The error rate is reported. We denote by $\dagger$ error rates obtained by previous works. The number of unlabeled images is $50000$ minus the number of labels. }
\label{CIFAR10COMP}
\end{centering}
\end{table*}

\begin{table*}[h!]

\resizebox{1\textwidth}{!}{  
\begin{centering}
\begin{tabular}{llc}
\begin{tabular}{|ccc|}
\hline
\multicolumn{3}{|c|}{\cellcolor[HTML]{D0D0D0} \textsc{CIFAR-100}}  \\ \hline
\multicolumn{1}{|c|}{} & \multicolumn{2}{c|}{\# \textsc{Labels}} \\  \cline{1-3}
\multicolumn{1}{|c|}{Method} &  4k & 10k  \\ \hline
\multicolumn{1}{|c|}{Fully Supervised}   & 55.59 $\pm$ 0.91  &  40.84 $\pm$ 0.34  \\\hline\hline
\multicolumn{1}{|c|}{LDPSSL $\dagger$~\cite{iscen2019label}}& 46.20 $\pm$ 0.76 & 38.43 $\pm$ 1.88 \\ 
\multicolumn{1}{|c|}{MT $\dagger$~\cite{tarvainen2017mean}}& 45.36 $\pm$ 0.49 & 36.08 $\pm$ 0.51 \\
\multicolumn{1}{|c|}{LDPSSL + MT $\dagger$~\cite{iscen2019label}}& \textbf{43.73 $\pm$ 0.20} & 35.92 $\pm$ 0.47 \\ \hline\hline
\multicolumn{1}{|c|}{CycleCluster}& 45.19 $\pm$ 0.34 \%  & 35.65 $\pm$ 0.50 \\ \hline
\multicolumn{1}{|c|}{CycleCluster+MT}& 44.34 $\pm$ 0.26   & \textbf{34.98 $\pm$ 0.38}\\ \hline
\end{tabular}
&
\begin{tabular}{|ccc|}
\hline
\multicolumn{3}{|c|}{\cellcolor[HTML]{D0D0D0} \textsc{Mini ImageNet}}                                         \\ \hline
\multicolumn{1}{|c|}{} & \multicolumn{2}{c|}{\# \textsc{Labels}} \\ \cline{1-3}
\multicolumn{1}{|c|}{Method} & 4k & 10k  \\ \hline
\multicolumn{1}{|c|}{Fully Supervised}   & 74.59 $\pm$ 0.90 \% &  60.17 $\pm$ 0.50  \\\hline\hline
\multicolumn{1}{|c|}{LDPSSL $\dagger$~\cite{iscen2019label}}& 70.29 $\pm$ 0.81 & 57.58 $\pm$ 1.47 \\ 
\multicolumn{1}{|c|}{MT $\dagger$~\cite{tarvainen2017mean}}& 72.51 $\pm$ 0.22 & 57.55 $\pm$ 1.11 \\
\multicolumn{1}{|c|}{LDPSSL + MT $\dagger$~\cite{iscen2019label}}& 72.78 $\pm$ 0.15 & 57.35 $\pm$ 1.66 \\ \hline\hline
\multicolumn{1}{|c|}{CycleCluster}& \textbf{69.12 $\pm$ 1.05} & \textbf{54.27 $\pm$ 0.71} \\ \hline
\multicolumn{1}{|c|}{CycleCluster+MT}& \textbf{ 63.30 $\pm$ 0.29 }  & \textbf{53.47 $\pm$ 0.17} \\ \hline

\end{tabular}

\end{tabular}

\end{centering}

}
\caption{Comparison with SSL methods on CIFAR-100 and Mini-ImageNet. The error rate is reported. We denote by $\dagger$ error rates obtained by previous works. For CIFAR-100 and Mini-ImageNet the number of clusters $K$ was set to the number of classes $C$ the number of unlabeled images is $50000$ minus the number of labels.}
\label{doublecomp}
\end{table*}

\section{Experiments}
In this section, we detail the datasets and evaluation protocol used to evaluate our proposed framework as well as provide implementation, parameter and training details. 

\subsection{Datasets Description and Evaluation Protocol}
We evaluate our approach using three benchmarking datasets: CIFAR-10 \cite{cifardata}, CIFAR-100 \cite{cifardata} and Mini-ImageNet \cite{oneshotMiniImage}. For CIFAR-10 experiments were performed using 500,1k, 2k and 4k labels whilst for CIFAR-100 and Mini-Imagenet experiments, were ran using  4k and 10k labels. \textbf{Evaluation Protocol.} For each dataset, we use the official partition. We use the error rate as the evaluation metric, over a range of label totals. As is standard practice in the area, we quote the mean error rate and standard deviation over five splits. For fair comparisons in the ablation study and comparisons, we use the suggested splits of~\cite{iscen2019label}.

\begin{table*}[t!]

\resizebox{1\textwidth}{!}{  
\begin{centering}
\begin{tabular}{llc}
\begin{tabular}{|ccc|}
\hline
\multicolumn{3}{|c|}{\cellcolor[HTML]{D0D0D0} \textsc{CIFAR-10}}  \\ \hline
\multicolumn{1}{|c|}{} & \multicolumn{2}{c|}{\# \textsc{Labels}} \\  \cline{1-3}
\multicolumn{1}{|c|}{Method} &  1k & 4k  \\ \hline
\multicolumn{1}{|c|}{Fully Supervised}   & 39.189 $\pm$ 0.91  &  40.84 $\pm$ 0.34  \\\hline\hline
\multicolumn{1}{|c|}{CycleCluster N-RA}& 14.76 $\pm$ 0.34 & 10.52 $\pm$ 0.45 \\ 
\multicolumn{1}{|c|}{CycleCluster RA}& 8.52 $\pm$ 0.29 & 6.58 $\pm$ 0.18 \\ \hline
\end{tabular}
&
\begin{tabular}{|ccc|}
\hline
\multicolumn{3}{|c|}{\cellcolor[HTML]{D0D0D0} \textsc{MiniImageNet}}  \\ \hline
\multicolumn{1}{|c|}{} & \multicolumn{2}{c|}{\# \textsc{Labels}} \\  \cline{1-3}
\multicolumn{1}{|c|}{Method} &  1k & 4k  \\ \hline
\multicolumn{1}{|c|}{Fully Supervised}   & 74.59 $\pm$ 0.90  &  60.17 $\pm$ 0.50  \\\hline\hline
\multicolumn{1}{|c|}{CycleCluster N-RA}& 69.12 $\pm$ 1.05 & 57.82 $\pm$ 1.01 \\ 
\multicolumn{1}{|c|}{CycleCluster RA}& 56.36 $\pm$ 0.49 & 45.40 $\pm$ 0.37 \\ \hline
\end{tabular}

\end{tabular}

\end{centering}

}
\caption{The effect of including strong augmentations in the form of one RandAugment \cite{cubuk2020randaugment} sample. The error rate is reported for CycleCluster without RandAugment (N-RA) and with RandAugment (RA). The experimental parameters used were the same as in the prior experiments. We report results for both CIFAR-10 and MiniImageNet and see a large increase in performance upon the inclusion of RandAugment.}
\label{augcomp}
\end{table*}

The goal of our work is to directly compare clustering regularisation against $\delta$ perturbation approaches. This comparison of techniques is obscured by the use of powerful data augmentation and optimisation tricks.  Therefore, we first compare our method against $\delta$ perturbation approaches in the absence of strong augmentation schemes. We compare our work against: Ladder Networks~\cite{rasmus2015semi},  VAT \cite{miyato2018virtual}, SSL-GAN~\cite{salimans2016improved}, TSSDL~\cite{shi2018transductive}, MT~\cite{tarvainen2017mean}, LPDSSL~\cite{iscen2019label} and ICT ~\cite{verma2019interpolation}. We also experiment with a combination of our approach and MT, when optimising $L_s(X_L,Y_L,\theta)$, and use the Mean Teacher code provided by the original Mean Teacher approach and use \cite{tarvainen2017mean}. Furthermore, we demonstrate that augmentation can be easily combined with our approach to boost performance and combine our approach with RandAugment \cite{cubuk2020randaugment}. In addition to this comparison, we perform ablation experiments relating to the implementation of clustering regularisation including its full removal.

\begin{table*}[t!]
\begin{centering}
\resizebox{0.9\textwidth}{!}{  
    \begin{tabular}{|ccccc|}
    \hline
    \multicolumn{5}{|c|}{\cellcolor[HTML]{D0D0D0} \textsc{CIFAR-10}}                                         \\ \hline
    \multicolumn{1}{|c|}{} & \multicolumn{4}{c|}{\# \textsc{Labels}} \\ \cline{1-5}
    \multicolumn{1}{|c|}{\textsc{Method}} & 500 & 1k & 2k & 4k   \\ \hline
    \multicolumn{1}{|c|}{Fully Supervised} & 48.93 $\pm$ 0.80 & 39.18 $\pm$ 0.88 & 28.23 $\pm$ 0.49 & 21.20 $\pm$ 0.46\\ \hline
    \multicolumn{1}{|c|}{Purely Graphical} & 32.21 $\pm$ 1.56  & 22.31 $\pm$ 0.78   &	15.63$\pm$0.45 & 12.63$\pm$0.32\\ \hline
    \multicolumn{1}{|c|}{\textcolor{blue}{LR}=0.05 \textcolor{red}{E}=180 \textcolor{black}{K}=10}   & \ 21.58 $\pm$ 1.73  &  15.86 $\pm$ 0.83 &  13.00 $\pm$ 0.30 & 10.73 $\pm$ 0.36   \\
    \multicolumn{1}{|c|}{\textcolor{blue}{LR}=0.05 \textcolor{red}{E}=180 \textcolor{black}{K}=100}   & 20.94 $\pm$ 2.19 & 15.52 $\pm$ 0.88 & 12.79 $\pm$ 0.35 & 10.79 $\pm$ 0.45 \\
    \multicolumn{1}{|c|}{\textcolor{blue}{LR}=0.05 \textcolor{red}{E}=180 \textcolor{black}{K}=300}   & 21.36$\pm$ 0.99 & 16.98 $\pm$ 0.90  & 13.43 $\pm$ 0.66 & 11.28 $\pm$ 0.39 \\ \hline
    \multicolumn{1}{|c|}{\textcolor{blue}{LR}=0.03 \textcolor{red}{E}=400 K=10}   & 23.83$\pm$ 2.78 & 16.42 $\pm$ 1.00 & 12.76 $\pm$ 0.64 & 10.79 $\pm$ 0.39 \\ 
    \multicolumn{1}{|c|}{\textcolor{blue}{LR}=0.03 \textcolor{red}{E}=400 \textcolor{black}{K}=100}  & \textbf{19.35 $\pm$ 2.52} &  \textbf{14.76$\pm$ 0.34}  & \textbf{ 12.11 $\pm$ 0.40 } & \textbf{10.52 $\pm$ 0.45}\\
    \hline
    \end{tabular}}
\caption{Ablation study on how changing the number of clusters $K$ effects the final classification accuracy on the CIFAR-10 dataset. \textcolor{blue}{LR} = Learning Rate, \textcolor{red}{E}~=~Epochs and \textcolor{black}{K}~=~Clusters}
\label{cifar10clus1}
\end{centering}
%}
\end{table*}

\begin{table*}[t!]
\begin{centering}
\resizebox{0.57\textwidth}{!}{
\begin{tabular}{|ccc|}
\hline
\multicolumn{3}{|c|}{\cellcolor[HTML]{D0D0D0} \textsc{CIFAR-100}}                                         \\ \hline
\multicolumn{1}{|c|}{} & \multicolumn{2}{c|}{\# \textsc{Labels}} \\ \cline{1-3}
\multicolumn{1}{|c|}{\textsc{Method}} &  4k & 10k   \\ \hline
\multicolumn{1}{|c|}{Fully Supervised} & 55.59 $\pm$ 0.91 \% &  40.84 $\pm$ 0.34\% \\ \hline
\multicolumn{1}{|c|}{Purely Graphical} & 47.30 $\pm$ 1.21 \% &  39.44 $\pm$ 0.64\% \\ \hline
\multicolumn{1}{|c|}{Clusters=100}   & 45.19 $\pm$ 0.34 \%  & \textbf{35.65 $\pm$ 0.52\%}  \\
\multicolumn{1}{|c|}{Clusters=300}  & \textbf{45.18 $\pm$ 0.49\%} & 35.72 $\pm$ 0.21\% \\ \hline
\end{tabular}
}
\caption{The effect of over-clustering on the CIFAR-100 dataset. Using $L_0 = 0.05$ and 180 epochs of training.}
\label{cifar100clus}
\end{centering}
\end{table*}

\subsection{Implementation Details and Training Scheme}

\textbf{Implementation.} Our approach is built using PyTorch and our experiments were ran on one Nvidia P100 GPU. \textbf{Deep Nets Architecture.} For the CIFAR-10 and CIFAR-100 dataset we used the "13-layer" network, that has been used in previous works~\cite{laine2016temporal}, as the feature extractor. For Mini-Imagenet we use the ResNet-18 architecture~\cite{He_2016_CVPR}.We add an $l_2$ normalisation layer before the fully connected layers and set the dropout rate to zero.

\textbf{Hyper-parameters.}  For all experiments we used stochastic gradient descent with cosine based annealing \cite{cosineSGD} with the following parameters: momentum = $0.9$ and weight decay $2 \times 10^{-4}$. For Mini-ImageNet and CIFAR-100 we train for $180$ epochs with $l_0=0.05$ and an annealing finishing point of $210$ epochs and for  CIFAR-10 we use a longer training length of $400$ epochs with $l_0=0.03$ and an annealing finishing point of $460$ epochs. We perform supervised initialisation on the initial labels for ten epochs.  On all datasets, clustering was done for $100$ iterations of the $k$-means algorithm. Before clustering the data was $L_2$ normed. For CIFAR-10 we use a batch size $B=100$ with $B_L=50,B_U=50$ whilst for CIFAR-$100$ and Mini-ImageNet. we use a batch size of $B=128$ with $B_L=88,B_U=40$ for 

\subsection{Results and Discussion}

In this section we present the experimental results generated from the previously outlined experiments.

\textbf{Method Comparison.} We compare our proposed framework against several different $\delta$-perturbation models which offer a wide variety of the $\delta$-perturbations used in the field. For the compared methods we use the code provided by the authors. \textbf{CIFAR-10} We present the comparison results for CIFAR-10 in Table \ref{CIFAR10COMP}. We see that all methods considered improve their performance with more labelled data. However, the performance of SSL-GAN is particularly poor relatively to the other methods, supporting the prior work that has suggested adversarial training leads to poor generalisation. We note that our approach is the best performing method on CIFAR-10, posting the best result for all label splits. Furthermore, we can see by comparing our approach to LPDSSL that the inclusion of clustering based regularisation to a graphical approach offers far greater performance at low label amounts than a pure graphical approach. \textbf{CIFAR-100} We present the results for CIFAR-100 in Table \ref{doublecomp}. We find that our approach again performs well producing the lowest error rate for 10k labels but slightly below the performance of LDPSSL+MT on 4K labels. Our approach improves with MT added, decreasing the error rate for both 4k and 10k labels. For CIFAR-10 we found the addition of MT did not change the error rate which we attribute to the simple nature of the CIFAR-10 dataset.

\textbf{Mini-ImageNet}: For Mini-ImageNet Table \ref{doublecomp} our method is by some margin the best method considered. Note that the addition of MT reduces the performance of LDPSSL whilst our approach improves upon the performance of LDPSSL and is considerably better. We would like to highlight the amazing performance that our method combined with MT has on the Mini-ImageNet dataset. CycleCluster + MT achieves error rates of $10.67$ and $5.81$ better than LDPSSL+MT for 4k and 10k labels respectively. The results on CIFAR-100 and Mini-Imagenet suggests that clustering regularisation maybe particularly suited to certain datasets over others.

As recent approaches such as \cite{sohn2020fixmatch,berthelot2019mixmatch} have shown, the inclusion of stronger augmentation techniques can leave to a dramatic performance increase in the semi-supervised setting. The inclusion of augmentation to our framework is trivial and can be done separately for both the clustering and classification loss. In our case we choose to add one sampled augmentation from RandAugment \cite{cubuk2020randaugment} in addition to our standard flip and crop augmentation whilst keeping all parameters the same. We give results on both CIFAR-10 and Mini-ImageNet for our augmented version in Table \ref{augcomp} and compare that to the baseline model. We see that the inclusion of data augmentation greatly increases the performance of CycleCluster across all datasets and label numbers, demonstrating that data augmentation can easily be combined with clustering based regularisation.

\subsection{Ablation Study}
For clustering methods, the number of clusters $K$ has to be provided as a prior parameter for most methods. Therefore there maybe a risk that choosing a bad value of $K$ could harm performance rather than help. Therefore we perform training using several different values of $K$ to assess the effect on the network, including the over-clustering case where we use more clusters than classes. We also consider a variant of our approach where the clustering regularisation is completely removed, which we name "Purely Graphical", to isolate its effect on the approach. These results are reported in Tables \ref{cifar10clus1} and \ref{cifar100clus}. Firstly, we see that, for all values of $K$, clustering based regularisation drastically reduces the error rate from the purely graphical model. On CIFAR-10 we see the benefits of clustering regularisation are particularly strong for small numbers of labeled points.  We found that in the CIFAR-10, with its large number of images per class, a small amount of over-clustering increases the classification performance but too much slightly decreased it. For CIFAR-100, we found that the performance increase was not dependent upon $K$ which suggests that the performance increase of clustering regularisation maybe heavily dataset dependent. The improvement in performance from using over-clustering regularisation is very robust to the value of $K$ and \textit{choosing the value of $K$ is not a major problem in this framework}.

In addition to this baseline we also provide another variation of our model in with we use RandAugment augmentation on both the clustering loss $L_C$ and the semi-supervised loss $L_W$. We use the same RandAugment implementation as the FixMatch approach \cite{sohn2020fixmatch}. We present results for this augmented version of CycleCluster on both CIFAR-10 and MiniImageNet.

\section{Conclusions}
In the field of SSL, the vast majority of recent approaches rely upon the \textit{low density separation assumption} to boost performance. The implementation of this assumption is usually done by demanding invariance with respect to perturbations of the data input. However, this local approach to consistency disregards the global structure of the data. Therefore, in this work we propose a novel regularisation for SSL classification based upon the direct implementation of the clustering assumption. We propose a novel framework, termed CycleCluster, which simultaneously uses self-supervised and semi-supervised learning making use of graph-based label propagation. Our experimental results demonstrate that our implementation of clustering regularisation can greatly improve model performance even on complex datasets such as Mini-Imagenet. Highlighting that clustering regularisation is a strong viable alternative for improved model generalisation.

\section{Supplementary Material}

In this section we provide supplementary material for our CycleCluster methods that proposes and explores cluster regularisation for semi-supervised image classification. This document is split in the following way. In Section I we detail optimisation choices and provide a full algorithm for training CycleCluster from start to finish.

\subsection{Optimisation Details}
For CycleCluster we iteratively move between generating cluster and class based pseudo-labels and optimising our cluster loss $L_C$ and our semi-supervised loss $L_W$. Referring to Algorithm \ref{trainingcyclecluster} we first initialise our model on the small amount of labelled data initally available. We then enter our main loop. We first extract a feature representation of the dataset and use $K$ means clustering to produce cluster pseudo-labels $\tilde{Y}$ and graphical propogation with $L_2$ Laplacian to produce the class-based pseudo-labels $\hat{Y}$ at the same point. We additionally compute entropy weights $\omega_i$ for each image and a class weight $\zeta_j$ for each class. We then sequentially optimise $L_C$ for one pass through the whole dataset and optimise $L_W$ for one pass through the unlabelled data. The class and cluster pseudo-labels are then updated and optimisation occurs again etc. We choose this sequentially optimisation rather than a joint optimisation as we found that the final classification accuracy was higher if the model was effectively fine tuned to the task at hand prior to pseudo-label generation.

For the algorithm Lines 2-8 relate to the supervised initialisation. Lines 9-30 cover the main optimisation loop with lines 10-15 covering the calculating of cluster and class pseudo-labels, lines 16-21 covering the creation of entropy and class weighting parameters and finally lines 22-29 covering the sequential optimisation of the clustering and semi-supervised loss.

\begin{algorithm}
	\caption{Training CycleCluster}
	\label{trainingcyclecluster}
	\begin{algorithmic}[1]
		\State \textbf{Input}  Dataset $Z$ with labeled samples $Z_l= \{ x_i, y_i \}_{i=1}^{n_l}$ with $C$ total classes and unlabeled samples $Z_u = \{ x_i \}_{i=n_l+1}^{n}$, Model $f_{\theta}$ of composite functions $z_{\theta} , g_{\theta}$
		\State \textbf{Parameters}: Number of epochs $E$, Batch size $b$, Labeled batch size $b_l$, Unlabeled batch size $b_u$. 
		
		\For { i = 1,2,....,100}
		\For { $j = 1,.., \floor{\frac{n_l}{b}}$} \Comment{Initial Supervised Baseline}
		    \State Batch $B_L = \{x_i,y_i\}_{i=1}^{b} \subset Z_l $
		    \State $\theta \leftarrow L_s = \frac{1}{b} \sum_{i=1}^{b} l_s(f_{\theta}(x_i),y_i)  $
		\EndFor    
		\EndFor
		\For { i = 1,..,E}
		\State $V = \{v_1,..,v_n\} = z_{\theta}(X)$ where $X = \{x_1,..,x_n\}$ \Comment{Extract Feature Embeddings}
		\State Perform $K$-means clustering and extract $\tilde{Y}$ 
		\State Construct Graph Matrix $W$
        \State Degree Normalisation $\mathcal{W} =  D^{\frac{-1}{2}}WD^{\frac{-1}{2}} $
        \State Propagate Information via $\mathcal{Q}(F)$
        \State $\hat{y_i} = \argmax F_i \text{  } \forall \text{  } n_l+1 \leq i \leq n$  
		\For{ $1 \leq i \leq n$}
		\State Calculate entropy weight $w_i := 1 - \frac{H(F_{i})}{log(C)}$ 
		\Comment{H being Shannon Entropy}
		\EndFor 
		\For { $1 \leq j \leq C$}
		\State Calculate class weight $\zeta_j = \left( \sum_{i=1}^{n_l} \mathbbm{1}_{y_i = j} + \sum_{i=n_l+1}^{n} \mathbbm{1}_{\hat{y}_i = j} \right)^{-1}$
		\EndFor
		\For { $i = 1,.., \floor{\frac{n}{b}}$} \Comment{Clustering  Regularisation}
		\State Batch $B_C = \{x_i,\tilde{y}_i\}_{i=1}^{b} \subset \{Z,\tilde{Y}\} $
		\State $\theta \leftarrow \frac{1}{b} \sum_{i=1}^{b} l_s(f_\theta(x_i),\tilde{y}_i)$
		\EndFor  
		\For { $i = 1,.., \floor{\frac{n-n_l}{b}}$} \Comment{Semi-Supervised Learning}
		\State Batch $B_L = \{x_i,y_i\}_{i=1}^{b_l} \subset \{Z_l\} $ , $B_U = \{x_i,\hat{y}_i\}_{i=1}^{b_u} \subset \{Z_u,\hat{Y}\} $
		\State $\theta =  \frac{1}{b_l} \sum_{i=1}^{b_l} \zeta_{y_i} l_s (f_{\theta}(x_i),\hat{y}_i)
        + \frac{1}{b_u}\sum_{i=1}^{b_u}\zeta_{\hat{y}_i} \omega_i l_s (f_{\theta}(x_i),\hat{y}_i)$
		\EndFor
		
		\EndFor
	\end{algorithmic} 
\end{algorithm}

\vspace{2cm}

\bibliographystyle{IEEEtran}
\bibliography{IEEEabrv,bibfile.bib}

\end{document}